\documentclass[10pt,twocolumn,letterpaper]{article}

\usepackage{iccv}
\usepackage{times}
\usepackage{epsfig}
\usepackage{graphicx}
\usepackage{amsmath}
\usepackage{amssymb}
\usepackage{multicol}
\usepackage{multirow}
\usepackage{bm}

\usepackage[pagebackref=true,breaklinks=true,letterpaper=true,colorlinks,bookmarks=false]{hyperref}

\iccvfinalcopy 

\def\iccvPaperID{2873} 

\ificcvfinal\fi

\begin{document}

\title{Simpler is Better: Few-shot Semantic Segmentation with Classifier Weight Transformer}

\author{Zhihe Lu$^{1,2}$ Sen He$^{1,2}$ Xiatian Zhu$^1$ Li Zhang$^3$ Yi-Zhe Song$^{1,2}$ Tao Xiang$^{1,2}$ \\
$^1$CVSSP, University of Surrey \\ 
$^2$iFlyTek-Surrey Joint Research Centre on Artificial Intelligence \\
$^3$School of Data Science, Fudan University \\
{\tt\small \{zhihe.lu, sen.he, xiantian.zhu, y.song, t.xiang\}@surrey.ac.uk, lizhangfd@fudan.edu.cn}
}

\maketitle
\ificcvfinal\thispagestyle{empty}\fi

\begin{abstract}
A few-shot semantic segmentation model is typically composed of a CNN encoder, a CNN decoder and a simple classifier (separating foreground and background pixels). Most existing methods meta-learn all three model components for fast adaptation to a new class. However, given that as few as a single support set image is available, effective model adaption of all three components to the new class is extremely challenging. In this work we propose to simplify the meta-learning task by focusing solely on the simplest component -- the classifier, whilst leaving the encoder and decoder to pre-training. We hypothesize that if we pre-train an off-the-shelf segmentation model over a set of diverse training classes with sufficient annotations, the encoder and decoder can capture rich discriminative features applicable for any unseen classes, rendering the subsequent meta-learning stage unnecessary. For the classifier meta-learning, we introduce a Classifier Weight Transformer (CWT) designed to dynamically adapt the support-set trained classifier's weights to each query image in an inductive way. 
Extensive experiments on two standard benchmarks show that despite its simplicity, our method outperforms the state-of-the-art alternatives, often by a large margin. Code is available on \href{https://github.com/zhiheLu/CWT-for-FSS}{https://github.com/zhiheLu/CWT-for-FSS}.
\end{abstract}

\vspace{-0.4cm}
\section{Introduction}

\begin{figure}[t]
    \centering
    \includegraphics[width=0.48\textwidth]{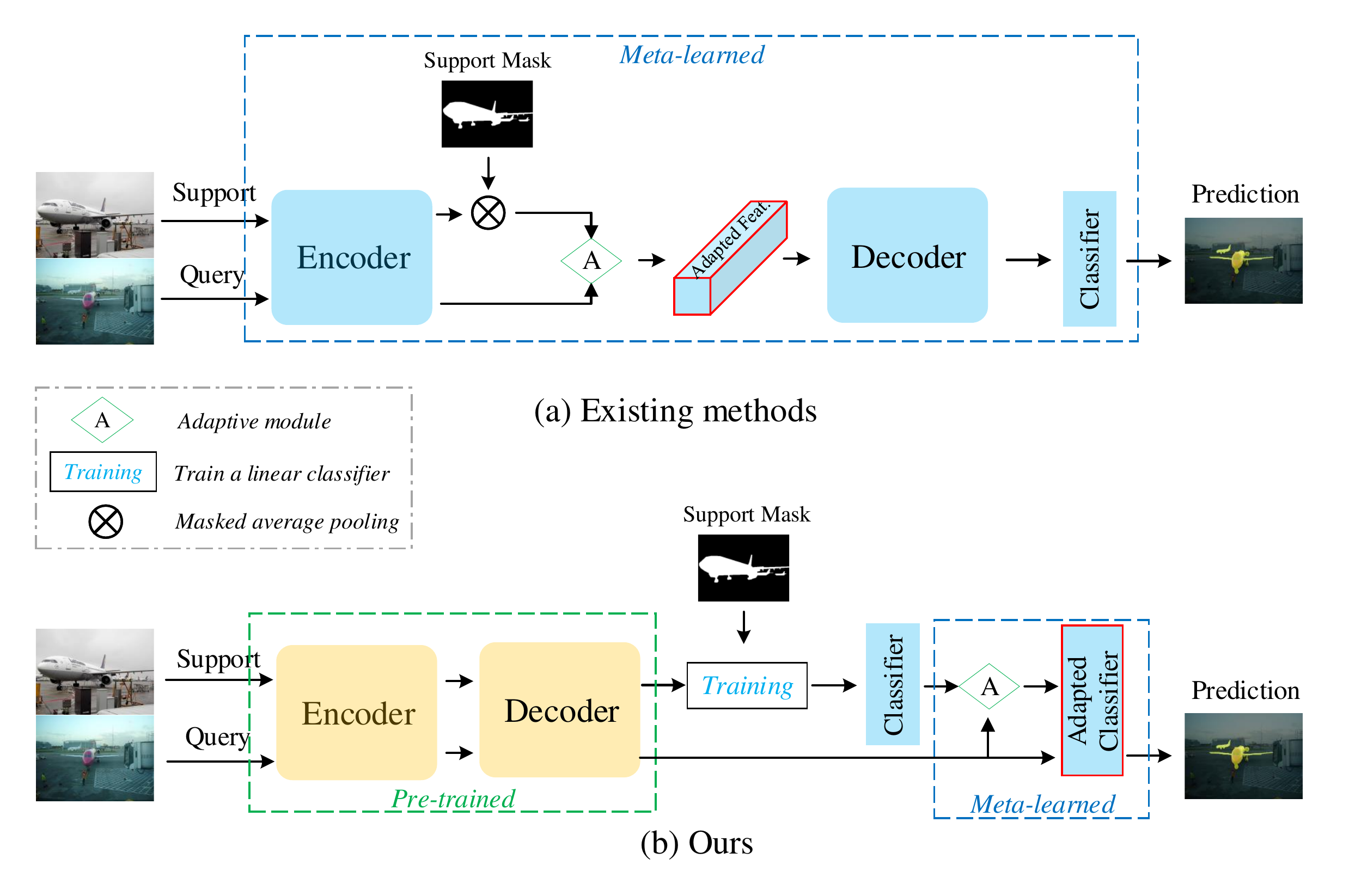}
    \caption{Illustrating the model training difference for 1-shot scenario between (a) existing few-shot segmentation methods and (b) ours. A few-shot segmentation model is typically composed of a deep CNN encoder, a deep CNN decoder and a much simpler classifier. (a) Previous training methods usually meta-learn all the three parts (\ie, the whole model), so that all three can adapt to a new class represented by an annotated support set image to perform segmentation on a query image. This adaptation is intrinsically difficult and sub-optimal due to complex model design and rather limited supervision available for the adaptation. 
    Instead, (b) we propose to meta-learn the simple classifier part only whilst pre-training and then freezing the encoder and decoder,
    making few-shot adaptation to new classes much more tractable.
    }
    \vspace{-0.4cm}
    \label{fig:motivation}
\end{figure}

Semantic segmentation has achieved remarkable progress in the past five years thanks to the availability of large-scale labeled datasets and advancements in deep learning algorithms~\cite{chen2017deeplab, chen2018encoder, long2015fully}. 
Nonetheless, relying on many training images with exhaustive pixel-level annotation for every single class, 
existing methods have poor scalability to new classes.  
Indeed, the high annotation cost
has hindered the general applicability of semantic segmentation models.  
For instance, creating the COCO dataset~\cite{lin2014microsoft} took over 70,000 worker hours
even for only 80 common object categories.
Inspired by the significant efforts in 
few-shot image classification \cite{snell2017prototypical,finn2017model,sung2018learning,ye2020few},
few-shot learning has been 
introduced into semantic segmentation recently
\cite{shaban2017one,nguyen2019feature, cao2020few, wang2019panet, yang2020prototype, zhang2019canet, zhang2020sg}.
A few-shot segmentation method eliminates the need of labeling a large set of training images. This is typically achieved by meta learning which enables the model to adapt to a new class represented by a support set consisting of as few as a single image. 



A fundamental challenge faced by any few-shot segmentation method is how to effectively adapt a complex image segmentation model to a new class represented by a small support set composed of only few images. More specifically, most recent segmentation models are deep CNNs with  complex architectures consisting of three parts: a CNN encoder, a CNN decoder and a classifier. Among the three, the classifier, which labels each pixel into either foreground classes or background, is much simpler compared to the other two -- in most cases, it is a $1 \times 1$ convolutional layer with the parameter number doubling the pixel feature dimension. 

As shown in Figure \ref{fig:motivation}(a), existing few-shot segmentation models~\cite{shaban2017one, yang2020prototype, zhang2019canet, zhang2020sg} aim to meta-learn all three parts. Concretely, existing few-shot segmentation methods mostly adopt an episodic training strategy. In each training episode, one training class is sampled with a small support set and query images to imitate the setting for testing.  Once learned, given a new class with a fully annotated support set and an unannotated query image, all three parts are expected to adapt to the new class so that foreground and background pixels can be separated accurately in the query image. 
Note that both the encoder and decoder are deep CNNs,
\eg, VGG-16 \cite{simonyan2014very} (15M parameters without fully-connected (FC) layers) or 
ResNet-50/101 \cite{he2016deep} (24M/43M parameters without FC layers) for encoder, and ASPP \cite{chen2017rethinking} (3.4M parameters) for decoder.
Effectively adapting the full model, especially the encoder and decoder, is thus a daunting task, hindering the performance of the existing models.

\begin{figure}
    \centering
    \includegraphics[width=0.45\textwidth]{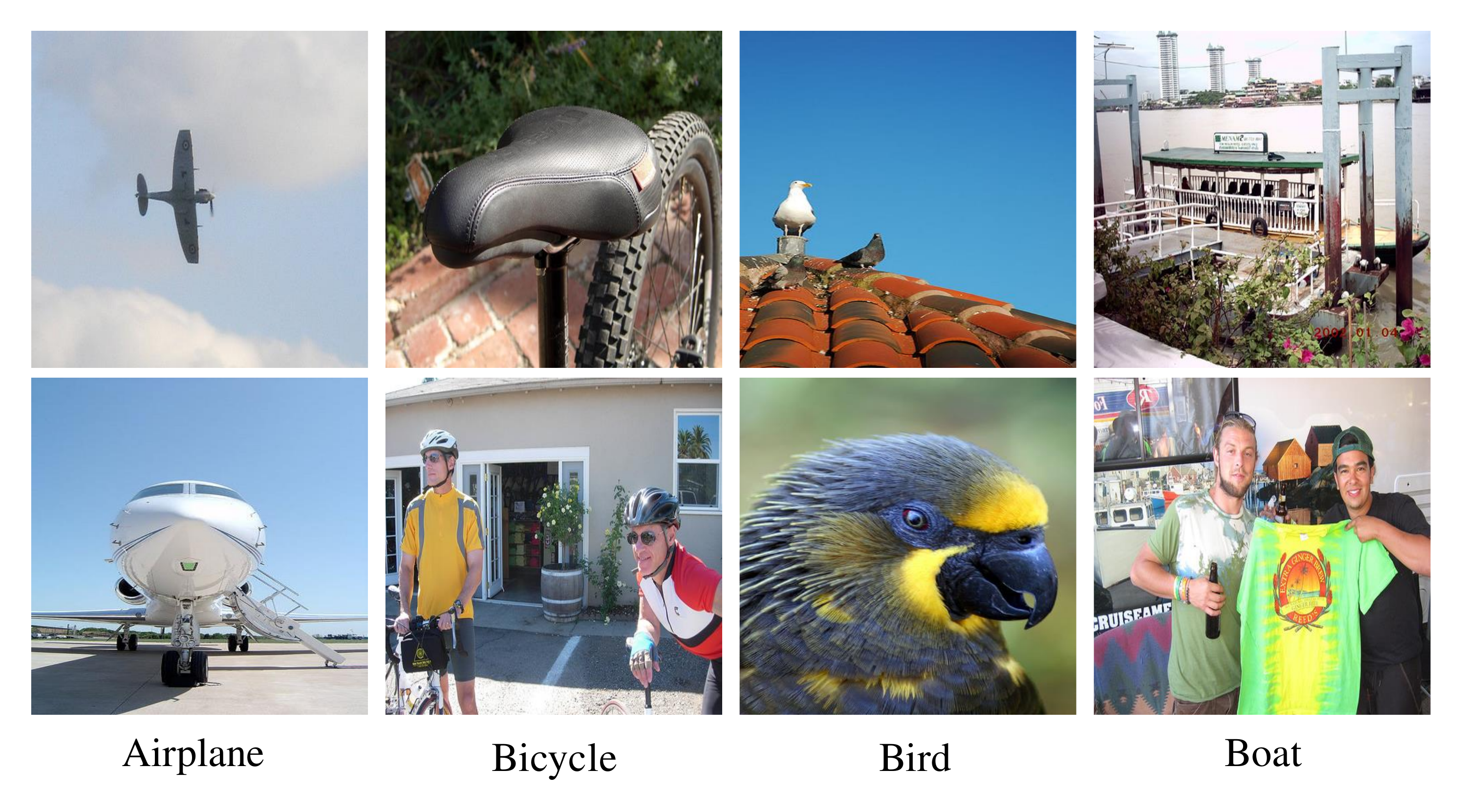}
    \caption{Examples of large intra-class variation in unconstrained images.
    It is evident that the objects from the same class (in each column) may look rather different due to uncontrolled changes in spatial size, viewpoint, style, and occlusion.
    }
    \vspace{-0.4cm}
    \label{fig:examples}
\end{figure}

%
%

To overcome the above fundamental limitations with existing few-shot segmentation methods, we conjecture that the key is to {\em simplify the meta-learning task} so that few-shot learning becomes more tractable and hence more effective. 
To this end,  we propose to focus meta-learning on the simplest and latest stage of the three-part pipeline -- {\em the classifier} whilst leaving the training of the encoder and decoder to a pre-training stage (see Figure \ref{fig:motivation}(b)). Once trained, only the classifier needs to be adapted to the new class with the rest of the model frozen, drastically reducing the complexity of model adaptation.  Our assumption is that if we pre-train an off-the-shelf segmentation model over a set of diverse training classes, the encoder and decoder can already capture a rich set of discriminative segmentation features suitable for not only the training classes but also the unseen test classes, \ie, being class-agnostic. We need to then focus on adapting the classifier part alone. Indeed, we found that if we simply do the pre-training and use the support set of a new class to train a classifier (\ie, without meta-learning), the result is already comparable to that obtained by the state-of-the-art methods (see Sec. \ref{sec:exp}). 
However, this naive baseline cannot adapt to each query image, which is critical for our problem due to the large intra-class variation (Figure \ref{fig:examples}). Without sufficient training samples in the support set to accommodate this intra-class variation, a few-shot segmentation model must adapt to each query image as well.   We hence further propose a novel meta-learning framework that employs a Classifier Weight Transformer (CWT) to dynamically adapt the support-set trained classifier's weights to each query image in an {\em inductive} way, \ie, adaptation occurs independently on each query image. 


We make the following contributions in this work:
(1) We propose a novel model training paradigm for few-shot semantic segmentation. Instead of meta-learning the whole, complex segmentation model, we focus on the simplest classifier part to make new-class adaptation more tractable.
(2) We introduce a novel meta-learning algorithm that leverages a Classifier Weight Transformer (CWT) for adapting dynamically the classifier weights to every query sample.
(3) Extensive experiments with two popular backbones (ResNet-50 and ResNet-101) show that the proposed method yields a new state-of-the-art performance, often surpassing existing alternatives, especially on 5-shot case, by a large margin. Further, under a more challenging yet practical cross-domain setting, the margin becomes even bigger.

\vspace{-0.1cm}
\section{Related Work}



\subsection{Few-shot Learning} 

Few-shot learning (FSL) aims to learn to learn a model for a novel task with only a handful of labeled samples. 
The majority of existing FSL works adopt the meta-learning paradigm~\cite{schmidhuber1987evolutionary}
and are mostly focused on image classification \cite{vinyals2016matching,finn2017model,snell2017prototypical,cai2018memory,sung2018learning,gidaris2018dynamic,liu2019prototype,su2020does,goldblum2020unraveling}.
As for which part of a classification model is meta-learned, this  varies in different works including
feature representation \cite{snell2017prototypical}
and distance metrics  \cite{sung2018learning}.
Beyond image classification, this learning paradigm can be applied to many different computer vision problems including semantic segmentation as investigated in this work.

%

\subsection{Few-shot Semantic Segmentation}
Recently, meta-learning has been introduced into semantic segmentation for addressing the same few-shot learning challenge \cite{shaban2017one}.
A semantic segmentation system generally consists of three parts: an encoder, a decoder and a classifier.
To incorporate meta-learning, 
a common strategy existing works consider involves two steps:
first relating the support-set and query-set image features from the encoder, and then updating all three parts by minimizing a loss measuring the discrepancy between the prediction and the ground-truth of query samples.
In terms of how to relate the support and query images,
there exist two different ways:
prototypical learning \cite{dong2018few, wang2019panet, liu2020part},
and feature concatenation~\cite{cao2020few, azad2021texture, zhang2019canet, yang2020prototype}.
PL~\cite{dong2018few} is the first work introducing prototypical learning into few-shot segmentation which predicts foreground/background classes by similarity comparison to prototypes. PANet~\cite{wang2019panet} further introduced a prototype alignment regularization to do bidirectional prototypical learning. 
Recently, PPNet~\cite{liu2020part} emphasized the importance of fine-grained features and proposed part-aware prototypes. 
By contrast, feature concatenation based methods first combine prototypes and query features, and then utilize a segmentation head, \eg,  ASPP~\cite{chen2017rethinking} and PPM~\cite{zhao2017pyramid}, for the final prediction. 
Despite the differences in model design, existing methods share a  common characteristic, \ie, they all attempt to update the whole complex model with just a few samples during meta-learning. 
This may cause optimizing difficulty as we mentioned before.
To overcome this issue, we propose to only meta-learn \textit{the classifier} during meta-training.

%

To address the intra-class variation challenge,
we further introduce a Classifier Weight Transformer (CWT)
to adapt the support-set trained classifier to every query image.
Our CWT is based on the self-attention mechanism \cite{vaswani2017attention}.
Motivated by the great success in NLP,
researchers have started to employ self-attention 
for vision tasks such as object detection \cite{hu2018relation,carion2020end}, and
image classification \cite{wu2020visual,dosovitskiy2020image}.
The closest work to ours is FEAT~\cite{ye2020few}
which leverages a prototype Transformer to calibrate 
the relationships of different classes 
for few-shot image classification.
However, in this work we explore the Transformer differently
for tackling the {\em intra-class} variation problem
in few-shot segmentation.

\section{Methodology}

\subsection{Task Definition}
We adopt the standard few-shot semantic segmentation setting \cite{shaban2017one,cao2020few}.
Given a meta-test dataset $D_\text{test}$,
we sample a target task with $K$-shot labeled images (\ie, the {\em support set}) and several test images (\ie, the {\em query set}) from one random class
and test a learned segmentation model $\bm{\theta}$.
The objective is to segment all the objects of the new class in the query images. 
To train the model $\bm{\theta}$ in a way that
it can perform well on those sampled segmentation tasks,
episodic training is adopted to meta-learn the model.
Concretely, a large number of such tasks are randomly sampled 
from a meta-training set $D_\text{train}$,
and then used to train the model in an episodic manner.

In each episode, we start with sampling one class $c$
from $D_\text{train}$ at random,
from which labeled training samples are then randomly drawn
to create a {\em support} set $\mathcal{S}$
and a {\em query} set $\mathcal{Q}$ with
$K$ and $Q$ samples, respectively.
Formally, the support and query sets are defined as:
\begin{align}
    \mathcal{S} = \{ (x_i, M_i) \}_{i=1}^{K}, \;\; 
    \mathcal{Q} = \{ (x_j, M_j) \}_{j=1}^{Q},
\end{align}
where $M_{i/j}$ denotes the ground-truth mask.
%
Note, $\mathcal{S} \cap \mathcal{Q} = \emptyset$ are sample-wise disjoint and $Q$ is 1 for the currently standard setting.

We conduct episodic training in a two-loop manner \cite{snell2017prototypical}:
the support set is first used in the inner loop to construct a classifier for the sampled class,
and the query set is then utilized in the outer loop to adapt the classifier with a Classifier Weight Transformer (CWT).

%
The key is to obtain a learner able to recognize any novel class with only a few labeled samples.
Compared with the standard segmentation setup, this is a more challenging task due to lacking sufficient supervision for new target classes. 
$D_\text{train}$ and $D_\text{test}$
contain base classes $C_{base}$ and novel classes $C_{novel}$, 
which are mutually disjoint, \ie,
$C_{base} \cap C_{novel} = \emptyset$.
Unlike the sparsely annotated meta-test classes, 
each meta-training class comes with abundant labeled training data so that sufficient episodes for model meta-training can be formed.



\begin{figure*}
    \centering
    \includegraphics[width=0.9\textwidth]{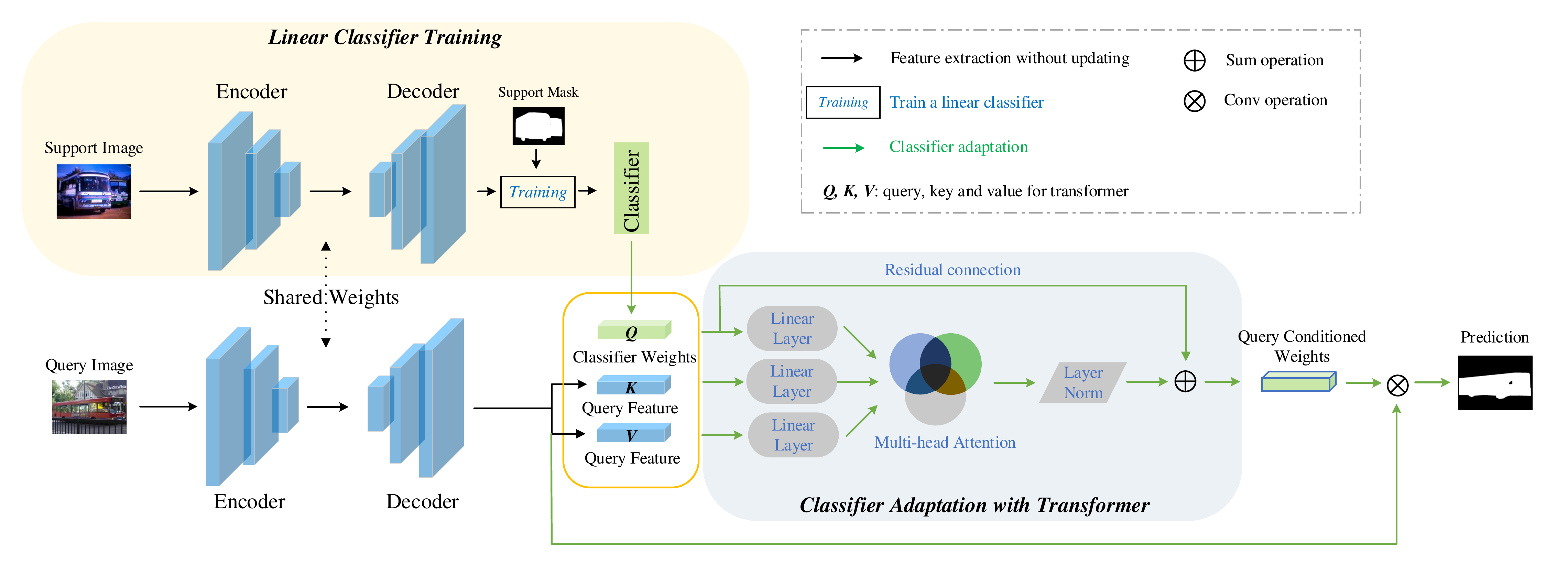}
    \caption{Schematic overview of the proposed few-shot semantic segmentation method. 
    Our model is trained in two stages.
    In the {\em first} stage,
    we pre-train the encoder and decoder on the base classes in a standard supervised learning manner.
    In the {\em second} stage, given an episode
    we froze the encoder and decoder, 
    initialize the classifier on the support set,
    and meta-learn a Classifier Weight Transformer (CWT) to udpate the classifier for each
    query image.
    During meta-testing, the classifier is first trained on the support set, then updated by the frozen CWT to adapt to any query image,
    and finally applied to  that query image for segmentation.
    }
    \vspace{-0.4cm}
    \label{fig:framework}
\end{figure*}

\subsection{Model Architecture}

A few-shot segmentation model generally consists of three modules: an encoder, a decoder and a classifier.
For learning to adapt to a new class, existing methods typically meta-learn the entire model after 
%
the encoder is pre-trained on ImageNet~\cite{russakovsky2015imagenet}. During the episodic training stage, all three parts of the model are meta-learned. Once trained, given a new class with annotated support set images and query images for test, the model is expected to adapt all three parts to the new class. With only few annotated support set images and the complex and interconnected three parts, this adaptation is often sub-optimal. 

To overcome these limitations we propose a simple yet effective training paradigm in two stages.
In the {\em first} stage, we pre-train the encoder and decoder for stronger feature representation with supervised learning.
In the {\em second} stage, together with the frozen encoder and decoder we meta-train the classifier only.
This is because we consider the pre-trained feature representation parts (\ie, the encoder and decoder) are sufficiently generalizable to any unseen classes;
the key for few-shot semantic segmentation would thus be in adapting the binary classifier (separating foreground and background pixels) rather than the entire model from few-shot samples.
The overview of our method is depicted in Figure \ref{fig:framework}.



\subsection{Stage 1: Model Pre-training}

As in all existing few-shot semantic segmentation models \cite{shaban2017one,cao2020few}, one of the key objectives is to learn the feature representation parts (\ie, the encoder and decoder) through meta learning, so that it can generalize to any unseen classes. 
%
For example, the state-of-the-art RPMMs model \cite{yang2020prototype}
was {\em directly} meta-trained with the encoder pre-trained on ImageNet.
However, in most recent few-shot learning methods for static image classification \cite{ye2020few,zhang2020deepemd}, pre-training the feature network on the whole meta-training set before episodic training starts has become a standard step. In this work, we also adopt such a pre-training step and show in our experiments that this step is vital (see Sec.~\ref{sec:ablation}).

Specifically, we use the PSPNet \cite{zhao2017pyramid} as our backbone segmentation  model. 
It is then pre-trained on the whole training set $D_\text{train}$ with the cross-entropy loss. 
Training details are provided in Sec \ref{sec:pretraining}.

\subsection{Stage 2: Classifier Meta-learning with Classifier Weight Transformer (CWT)}

After the pre-training stage, the encoder and decoder can be simply frozen for any different few-shot tasks.
Since any new task involves a previously unseen class,
the classifier has to be learned.
To this end, an intuitive and straightforward method is to
optimize the classifier weights $\bm{w}$ with the labeled support set $\mathcal{S}$.
With the pre-trained encoder and decoder we first extract the feature vectors $\bm{f} \in \mathcal{R}^d$ for every support-set image pixel.
The feature dimension is denoted as $d$.
Same as in pre-training, we then adopt the cross-entropy loss function to train the classifier model $\bm{w}$.

As the feature representation is considered to be class generic,
they can be used directly with the newly trained classifier for meta-test without going through the episodic training process. 
Especially, after seeing sufficient diverse classes, it can work well when the task is to separate foreground and background pixels.
Indeed, our experiments show that this turns out to be a surprisingly strong baseline that even outperforms a state-of-the-art PPNet \cite{liu2020part} (see Tables \ref{tab:main_coco} and \ref{tab:baseline}).
This verifies for the first time that good feature representation (or feature reuse) is similarly critical for few-shot semantic segmentation modeling -- a finding that has been reported in recent static image few-shot learning works  \cite{tian2020rethinking,liu2020negative}.

Nonetheless, this baseline is still limited for few-shot segmentation
since it cannot adapt to every query image whereby the target object may appear drastically dissimilar to the ones in the support-set images.
To that end, we next introduce our meta-learning algorithm that learns a Classifier Weight Transformer (CWT) for query object adaptation.





During episodic training, 
we aim to learn via our CWT  how to  adapt the classifier weights $\bm{w} \in \mathbb{R}^{2 \times d}$ to a sampled class in each episode.
%
%
Formally, the input to our transformer 
is in the triplet form of
(\texttt{Query}, \texttt{Key}, \texttt{Value}). 
We start with extracting the feature $\bm{F} \in \mathbb{R}^{n \times d}$ for all $n$ pixels of the query image using the encoder and decoder. 
To learn discriminative query conditioned information,
the input is designed as:
\begin{align}
    \text{Query} = \bm{w} \textbf{W}_q, \;\;  
    \text{Key} = \bm{F} \textbf{W}_k, \;\;
    \text{Value} = \bm{F} \textbf{W}_v,
\end{align}
where $\textbf{W}_q$/$\textbf{W}_k$/$\textbf{W}_v  \in \mathbb{R}^{d\times d_a}$
are the learnable parameters 
(each represented by a fully connected layer) 
that project the classifier weights and query feature to a $d_a$-D latent space.
To adapt the classifier weights to the query image,
we form a {\em classifier-to-query-image attention} mechanism 
as:
\begin{equation}
    \textbf{w}^* = \textbf{w} + 
    \psi (\operatorname{softmax}(\frac{\bm{w} \textbf{W}_q (\bm{F} \textbf{W}_k)^\top}{\sqrt{d_a}}) (\bm{F} \textbf{W}_v)),
    \label{eq:attn}
\end{equation}
where $\operatorname{softmax}(\cdot)$ is a row-wise softmax function for attention normalization and $\psi(\cdot)$ is a linear layer with the input dimension $d_a$ and output dimension $d$.
Residual learning is adopted for more stable model convergence.

As written in Eq.~\eqref{eq:attn}, pairwise similarity 
defines the attentive scores between the classifier weight and query image pixels,
and is further used for weighted aggregation in the \texttt{Value} space.
This adapts the classifier weight to the query image.
The intuition is that, the pairs involving a query image pixel from the new class often enjoy higher similarity than those with background classes except few outlier instances; as a result, this attentive learning would reinforce this desired proximity and adjust the classifier weights conditioned on the query.
Consequently, the intra-class variation can be mitigated.

{\bf Learning objective }
Once the classifier weight $\bm{w}^*$ is adapted to a query sample,
we then apply it for segmentation prediction on the corresponding feature
$\bm{F}$.
To train our transformer,
a cross-entropy loss can be derived from the query-image ground-truth and the prediction as the meta-training objective.

{\bf Unseen class adaptation }
During meta-testing, given any new task/class this proposed transformer can be directly applied 
to condition the classifier's weights first optimized on the support set to any given query image. Note that both support set and query images are used as input to our transformer to update the foreground/background classifier for the new unseen foreground class. However, the transformer parameter is fixed after meta-training and the adaptation is done for each query image independently, \ie, in an inductive manner as in most existing few-shot segmentation works. 


\section{Experiments}
\label{sec:exp}

\subsection{Datasets and Settings}
In our experiments, two standard few-shot semantic segmentation datasets are used.
\vspace{-0.4cm}

\paragraph{COCO-$20^{i}$} is currently the largest and most challenging dataset for few-shot semantic segmentation. 
It provides the train/val sets including 82,081/40,137 images in 80 classes, built from the popular COCO~\cite{lin2014microsoft} benchmark. Following~\cite{nguyen2019feature} we divide the 80 classes into 4 splits $i \in \{0,1,2,3\}$, each of which contains 20 classes. In a single experiment, three class splits are selected as the base classes for training whilst the remaining class split for testing. 
Therefore a total of four experiments are conducted.

\vspace{-0.4cm}


\paragraph{PASCAL-$5^{i}$} is the extension of PASCAL VOC 2012~\cite{everingham2010pascal} with extra annotations from SDS dataset~\cite{hariharan2014simultaneous}. The train and val sets contain 5,953 and 1,449 images, respectively. There are 20 categories in both the training set and the validation set. 
Following~\cite{shaban2017one} we make 4 class splits each with 5 classes and design the experiments in a similar protocol as COCO-$20^{i}$.

\begin{table*}[h]
    \centering
    \begin{tabular}{c|c|ccccc|ccccc}
    \hline
    \hline
         \multirow{2}{*}{Backbone} & \multirow{2}{*}{Methods} & \multicolumn{5}{c}{1-shot} & \multicolumn{5}{c}{5-shot}  \\
         \cline{3-12}
          & & s-0 & s-1 & s-2 & s-3 & Mean & s-0 & s-1 & s-2 & s-3 & Mean \\
         \hline
         \multirow{4}{*}{ResNet-50} & PANet~\cite{wang2019panet} (ICCV19)$^{\dag}$ & 31.5 & 22.6 & 21.5 & 16.2 & 23.0 & 45.9 & 29.2 & 30.6 & 29.6 & 33.8 \\
         & RPMMs~\cite{yang2020prototype} (ECCV20) & 29.5 & \textbf{36.8} & 29.0 & 27.0 & 30.6 & 33.8 & 42.0 & 33.0 & 33.3 & 35.5 \\
         & PPNet~\cite{liu2020part} (ECCV20)
         & \textbf{34.5} & 25.4 & 24.3 & 18.6 & 25.7 & \textbf{48.3} & 30.9 & 35.7 & 30.2 & 36.2 \\
         & CWT (Ours) & 32.2 & 36.0 & \textbf{31.6} & \textbf{31.6} & \textbf{32.9} & 40.1 & \textbf{43.8} & \textbf{39.0} & \textbf{42.4} & \textbf{41.3}  \\
         \hline
         \multirow{3}{*}{ResNet-101} & FWB~\cite{nguyen2019feature} (ICCV19) & 19.9 & 18.0 & 21.0 & 28.9 & 21.2 & 19.1 & 21.5 & 23.9 & 30.1 & 23.7 \\
         & PFENet~\cite{tian2020prior} (TPAMI20) & \textbf{34.3} & 33.0 & \textbf{32.3} & 30.1 & \textbf{32.4} & \textbf{38.5} & 38.6 & 38.2 & 34.3 & 37.4 \\
         & CWT (Ours) & 30.3 & \textbf{36.6} & 30.5 & \textbf{32.2} & \textbf{32.4} & \textbf{38.5} & \textbf{46.7} & \textbf{39.4} & \textbf{43.2} & \textbf{42.0}  \\
         \hline
         \hline
    \end{tabular}
    \vspace{2pt}
    \caption{Few-shot semantic segmentation results on COCO-$20^{i}$. 
    For a fair comparison among all methods,
    we compare with the results of PPNet~\cite{liu2020part}
    without extra unlabeled training data. $\dag$: The results cited from PPNet~\cite{liu2020part}. 
    }
    \label{tab:main_coco}
\end{table*}

\subsection{Implementation Details and Metrics}

\paragraph{Pre-training} 
\label{sec:pretraining}
To obtain a strong encoder and decoder,
we adopt the standard supervised learning for semantic segmentation on base classes, \ie, 16/61 base classes (including the background class) in each split of PASCAL-$5^{i}$/COCO-$20^{i}$. We choose PSPNet~\cite{zhao2017pyramid} as our baseline segmentation model. 
For fair comparisons with existing methods, we tested two common backbones, ResNet-50 and ResNet-101~\cite{he2016deep}. 
We also present the results with VGG-16 in the supplementary.
We train a model for 100 epochs on PASCAL-$5^{i}$ and 20 epochs on COCO-$20^{i}$.
We set the batch size to 12, the image size to 417.
The objective function is the cross-entropy loss. 
We use the SGD optimizer with the momentum 0.9, 
the weight decay 1e-4, the initial learning rate 2.5e-3,
and the cosine learning rate scheduler. 
We set the label smoothing parameter $\epsilon$ to 0.1.
For data augmentation, we only use random horizontal flipping.
\vspace{-0.5cm}


\paragraph{Episodic Training}
After pre-training, we froze the encoder and decoder in the subsequent episodic training. 
In this stage, we form the training data of base classes into episodes, each including a support set and a query set from a randomly selected class. 
We first train a new classifier for the selected class for 
200 iterations on the support set using the SGD optimizer and cross-entropy loss function.
The learning rate 1e-1 is used.
Next, we train the proposed CWT once at learning rate of 1e-3.
There are total 20 epochs for above inner- and outer- loop optimization.
Our transformer has a shared linear layer with dimension, $512 \times 2048$, for projecting the inputs to latent space, a 4-head attention module and a fully connected layer recovering the dimension to 512. In addition, a dropout layer for stable training and layer normalization are used.
It outputs query-adaptive classifier weights
which will be used to predict every pixel of the query images.
In meta-learning setup, the classifier resides in the inner loop whilst the transformer is in the outer loop.
\vspace{-0.5cm}



\paragraph{Evaluation Metrics}
We use the class mean intersection over union (mIoU) as the evaluation metric. 
Specifically, mIoU is computed over averaging the IoU rates of each class. Following~\cite{liu2020part}, we report all the results averaged across 5 trials. For each trial, we test 1,000 episodes. 
We report the results for every single split and their average.

\subsection{Single Domain Evaluation}

\subsubsection{COCO-$20^{i}$ Results}

In Table~\ref{tab:main_coco} we compare the segmentation results of our method and the latest state-of-the-art models on COCO-$20^{i}$.
We consider 1-shot and 5-shot cases, and two backbone networks (ResNet-50 and ResNet-101) for a more extensive comparison. 
Overall, the performance advantage of our method over all competitors
is significant.
For example, in 1-shot case we obtain 2.3\% mIoU gain over the best competitor with ResNet-50 backbone.
Much more gains (5.1\%/4.6\% with ResNet-50/ResNet-101) are shown in 5-shot case.
Similar to PPNet, our method can benefit consistently
from support set expansion.
In contrast, some existing methods (\eg, RPMMs, FWB and PFENet) are 
clearly inferior in leveraging extra labeled samples.
More importantly, all the compared SOTA baselines attempt to adapt all three parts of a segmentation model to the new class and each query image. Nevertheless, our method is much simpler by focusing on the linear classifier only. The superior results achieved by our method thus verify our assumption that the pre-trained encoder and decoder is generalizable; and meta-learning and adaptation are thus only necessary for the classifier.
Furthermore, we investigate the inference speed under a challenging yet practical scenario, \ie, 1,000 query samples per task on 1-shot case, as a trained model is expected to serve more images.
Our model runs at 21.7 frame per second (FPS) \textit{vs.} 18.9 FPS achieved by RPMMs~\cite{yang2020prototype} with ResNet-50.

\subsubsection{PASCAL-$5^{i}$ Results}

In Table~\ref{tab:main_pascal} we show the comparative results on PASCAL-$5^{i}$. 
On this less challenging dataset with a smaller number of object 
classes, we have similar observations as on COCO-20$^i$.
Our proposed method again achieves the best overall performance.
It is also noted that our model's advantage over the competitors is less pronounced compared to the COCO results.
Our method even performs worse in 1-shot case.
A plausible reason is that with far less training classes and images on PASCAL-$5^{i}$, our assumption of the pre-trained encoder/decoder being class-agnostic is not valid anymore. Existing methods' approach of adapting them to new class thus has some benefit contributing to narrowing down the gap to our method. 

\begin{table*}[]
    \centering
    \begin{tabular}{c|c|ccccc|ccccc}
    \hline
    \hline
         \multirow{2}{*}{Backbone} & \multirow{2}{*}{Methods} & \multicolumn{5}{c}{1-shot} & \multicolumn{5}{c}{5-shot} \\
         \cline{3-12}
          & & s-0 & s-1 & s-2 & s-3 & Mean & s-0 & s-1 & s-2 & s-3 & Mean \\
         \hline
         \multirow{6}{*}{ResNet-50} & CANet~\cite{zhang2019canet} (CVPR19) & 52.5 & 65.9 & 51.3 & 51.9 & 55.4 & 55.5 & 67.8 & 51.9 & 53.2 & 57.1 \\
         & PGNet~\cite{zhang2019pyramid} (ICCV19) & 56.0 & 66.9 & 50.6 & 50.4 & 56.0 & 57.7 & 68.7 & 52.9 & 54.6 & 58.5 \\
         & RPMMs~\cite{yang2020prototype} (ECCV20) & 55.2 & 66.9 & 52.6 & 50.7 & 56.3 & 56.3 & 67.3 & 54.5 & 51.0 & 57.3 \\
         & PPNet~\cite{liu2020part} (ECCV20) & 47.8 & 58.8 & 53.8 & 45.6 & 51.5 & 58.4 & 67.8 & 64.9 & 56.7 & 62.0 \\
         & PFENet~\cite{tian2020prior} (TPAMI20) & \textbf{61.7} & \textbf{69.5} & 55.4 & \textbf{56.3} & \textbf{60.8} & \textbf{63.1} & \textbf{70.7} & 55.8 & \textbf{57.9} & 61.9 \\
         & CWT (Ours) & 56.3 & 62.0 & \textbf{59.9} & 47.2 & 56.4 & 61.3 & 68.5 & \textbf{68.5} & 56.6 & \textbf{63.7} \\
         \hline
         \multirow{4}{*}{ResNet-101} & FWB~\cite{nguyen2019feature} (ICCV19) & 51.3 & 64.5 & 56.7 & 52.2 & 56.2 & 54.9 & 67.4 & 62.2 & 55.3 & 59.9 \\
         & DAN~\cite{cao2020few} (ECCV20) & 54.7 & 68.6 & 57.8 & 51.6 & 58.2 & 57.9 & 69.0 & 60.1 & 54.9 & 60.5 \\
         & PFENet~\cite{tian2020prior} (TPAMI20) & \textbf{60.5} & \textbf{69.4} & 54.4 & \textbf{55.9} & \textbf{60.1} & \textbf{62.8} & \textbf{70.4} & 54.9 & \textbf{57.6} & 61.4 \\
         & CWT (Ours) & 56.9 & 65.2 & \textbf{61.2} & 48.8 & 58.0 & 62.6 & 70.2 & \textbf{68.8} & 57.2 & \textbf{64.7} \\
         \hline
         \hline
    \end{tabular}
    \vspace{2pt}
    \caption{Few-shot semantic segmentation results on PASCAL-$5^{i}$. 
    For a fair comparison among all methods,
    we compare with the results of PPNet~\cite{liu2020part}
    without extra unlabeled training data.
    }
    \label{tab:main_pascal}
\end{table*}

\begin{table*}[h]
\small
    \centering
    \begin{tabular}{c|c|ccccc|ccccc}
    \hline
    \hline
         \multirow{2}{*}{Setting} & \multirow{2}{*}{Methods} & \multicolumn{5}{c}{1-shot} & \multicolumn{5}{c}{5-shot}  \\
         \cline{3-12}
          & & s-0 & s-1 & s-2 & s-3 & Mean & s-0 & s-1 & s-2 & s-3 & Mean \\
         \hline
         \multirow{2}{*}{COCO $\rightarrow$ PASCAL} & RPMMs~\cite{yang2020prototype} (ECCV20) & 36.3 & 55.0 & 52.5 & 54.6 & 49.6 & 40.2 & 58.0 & 55.2 & 61.8 & 53.8 \\
         & CWT (Ours) & \textbf{53.5} & \textbf{59.2} & \textbf{60.2} & \textbf{64.9} & \textbf{59.5} & \textbf{60.3} & \textbf{65.8} & \textbf{67.1} & \textbf{72.8} & \textbf{66.5}  \\
         \hline
         \multirow{2}{*}{PASCAL $\rightarrow$ COCO} & RPMMs~\cite{yang2020prototype} (ECCV20) & 27.0 & \textbf{44.7} & 40.6 & 33.2 & 36.4 & 30.2 & 47.8 & 46.2 & 39.6 & 41.0 \\
         & CWT (Ours) & \textbf{34.3} & 42.8 & \textbf{44.8} & \textbf{34.7} & \textbf{39.2} & \textbf{40.6} & \textbf{48.6} & \textbf{51.9} & \textbf{41.9} & \textbf{45.8} \\
         \hline
         \hline
    \end{tabular}
    \vspace{2pt}
    \caption{Few-shot semantic segmentation results
    in bidirectional cross-domain setting.
    Backbone: ResNet-50.
    }
    \label{tab:main_cross}
\end{table*}

\subsection{Cross Domain Evaluation}
Beyond the standard single domain few-shot segmentation setting,
we further introduce a more challenging and more realistic cross-domain setting.
In this new setting, we aim to test the generalization of a pre-trained model across previously unseen domains (datasets) with different data distributions.
This setting is more difficult yet more practical -- in real-world applications, the new segmentation tasks often involve both new classes and new domains.

In this experiment, we train a few-shot segmentation model on COCO-$20^{i}$/PASCAL-$5^{i}$ and then directly apply it to PASCAL-$5^{i}$/COCO-$20^{i}$ without any domain-specific model re-training or fine-tuning.
This data transfer design is a good cross-domain test, as the two datasets present a clear domain shift in terms of instance size, instance number and categories per image~\cite{lin2014microsoft}.
Concretely, from COCO to PASCAL, we use the original COCO-$20^{i}$ training class splits
for model training.
During testing on PASCAL-$5^{i}$, we take all the classes in validation set as a whole and remove the training classes seen in each split of COCO-$20^{i}$
to ensure the few-shot learning nature. 
For PASCAL-to-COCO case, the only difference from the original PASCAL-$5^{i}$ experiments is that the testing splits are from COCO that contains all the classes of PASCAL.

For comparative evaluation, we select the latest state-of-the-art method RPMMs~\cite{yang2020prototype}.
We use the models released by the RPMMs' authors for achieving its optimal results. 
We adopt the ResNet-50 backbone.
It is observed in Table~\ref{tab:main_cross} that our method is significantly superior on almost all the splits. For COCO-to-PASCAL setting, on average, our model yields a gain of 9.9\% and 12.7 \% in 1-shot and 5-shot cases, respectively.
From PASCAL to COCO, the improvements are 2.8\%/4.8\% for 1-shot/5-shot.
This suggests that our training method is more effective 
than conventional whole pipeline meta-learning for solving
the domain shift problem.
This is due to stronger feature representation
and a simpler and more effective learning to learn capability with focus on 
classifier adaptation.


\subsection{Ablation Study}
\label{sec:ablation}

We conduct a set of ablation experiments 
on the more challenging dataset COCO-20$^i$.
We take ResNet-50 as backbone and test the 1-shot case.

\begin{table}[]
    \centering
    \begin{tabular}{c|ccccc}
    \hline
         \multirow{2}{*}{Meta-training} & \multicolumn{5}{c}{1-shot}  \\
         \cline{2-6}
          & s-0 & s-1 & s-2 & s-3 & Mean \\
         \hline
         Whole model & 22.0 & 25.6 & 19.5 & 19.0 & 21.5 \\
         Classifier (Ours) & \textbf{32.2} & \textbf{36.0} & \textbf{31.6} & \textbf{31.6} & \textbf{32.9} \\
         \hline
    \end{tabular}
    \vspace{2pt}
    \caption{
    Meta-training the entire model {\em vs.} meta-training the classifier.
    Backbone: ResNet-50,
    Dataset: COCO-20$^i$.
    }
    \label{tab:whole_model}
\end{table}

\vspace{-0.3cm}

\subsubsection{What should be meta-learned, the whole model {\em or} the classifier only?}
\label{evidence}
A fundamental question we investigate in this paper is {\em what should be meta-learned} for few-shot semantic segmentation.
Existing methods usually meta-learn the whole model \cite{yang2020prototype, cao2020few},
which we consider to be sub-optimal  due to the number of model parameters to be updated given  very scarce training data.
Instead, we propose to only meta-learn the classifier.
For an exact comparison, we create a variant of our method 
which updates the whole model (\ie, the encoder, decoder, and classifier) in episodic training.
To enable one-stage training as our main model,
we compute a pair of prototypes for foreground and background classes from the support set 
and use it as the classifier's weights.
It is shown in Table~\ref{tab:whole_model} that
meta-learning the whole model is significantly inferior
by {11.4 \%} in mIoU.
This verifies our assumption that the meta-learning and adaptation of the whole complex segmentation model is neither effective nor necessary.

\begin{table}[h]
    \centering
    \begin{tabular}{c|ccccc}
    \hline
         \multirow{2}{*}{Component} & \multicolumn{5}{c}{1-shot}  \\
         \cline{2-6}
          & s-0 & s-1 & s-2 & s-3 & Mean \\
         \hline
         Classifier Only & 27.3 & 30.7 & 27.6 & 28.9 & 28.6  \\
         Classifier Adapt. & \textbf{32.2} & \textbf{36.0} & \textbf{31.6} & \textbf{31.6} & \textbf{32.9} \\
         \hline
    \end{tabular}
    \vspace{2pt}
    \caption{
    Model component analysis.
    Backbone: ResNet-50,
    Dataset: COCO-20$^i$.
    }
    \label{tab:baseline}
\end{table}

\vspace{-0.6cm}

\subsubsection{Component analysis}
\begin{figure}
    \centering
    \includegraphics[width=0.45
    \textwidth]{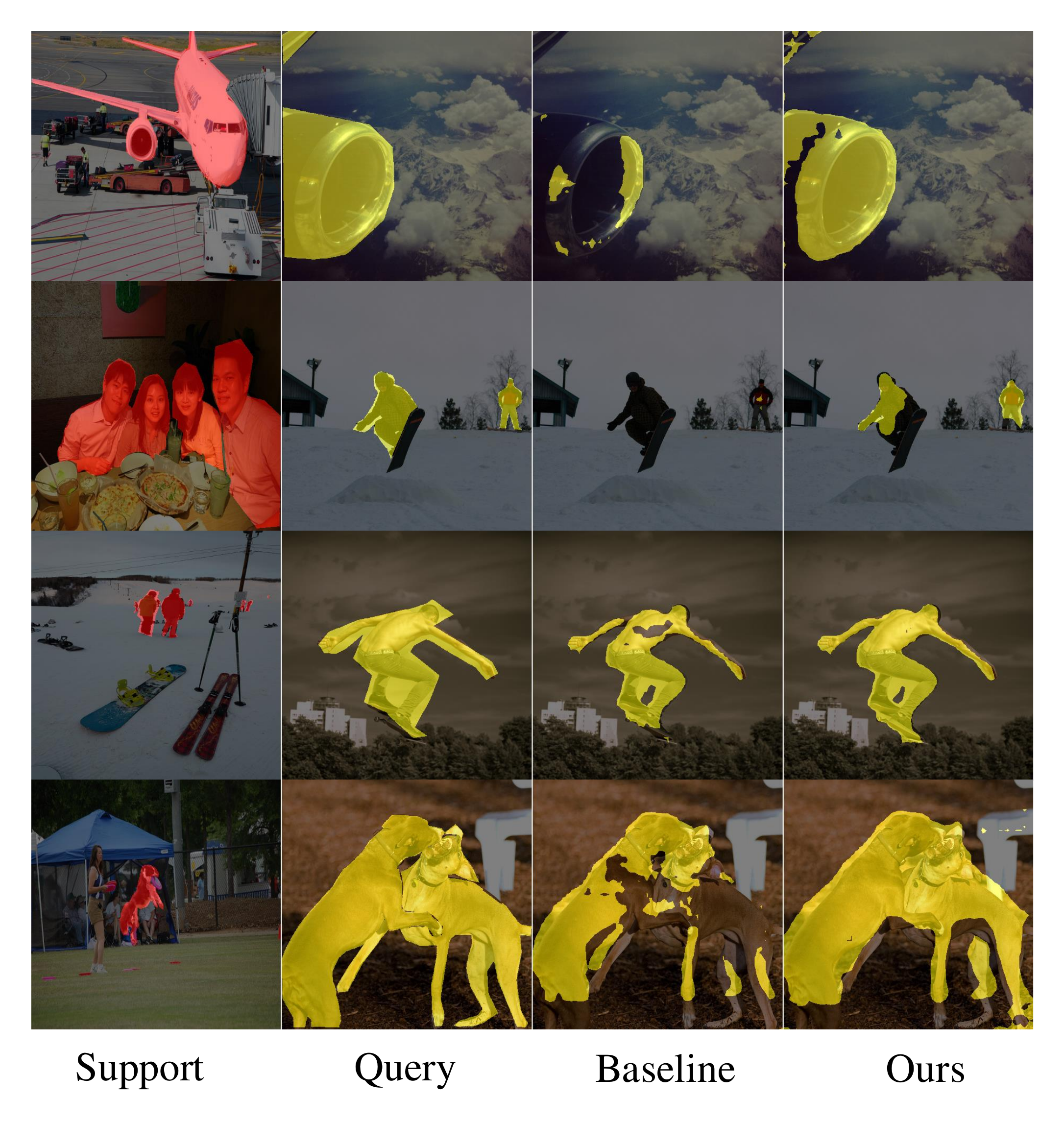}
    \caption{Quality results under 1-shot segmentation on COCO-$20^{i}$. From left to right, 
    support images with masks, 
    query images with masks, 
    baseline results, and our results.}
    \label{fig:visual}
\end{figure}

Our method can be decomposed into two phases:
model pre-training, and classifier adaptation.
Specifically, we pre-train the encoder and decoder (\ie, the feature representation components) on the whole training set.
To validate its effectiveness, 
we create a baseline of pre-training only without any meta-learning. That is, after pre-training, we go straight into testing and  
train a classifier on the support set for every meta-test task whilst freezing
the encoder and decoder.
The results in Table ~\ref{tab:baseline} reveal that
this turns out to be a very strong baseline.
For example, it even outperforms the latest model PPNet~\cite{liu2020part} by {2.9\%} (Table \ref{tab:main_coco}). 
This verifies the importance of model pre-training
for obtaining a strong class-agnostic feature representation,
and the efficacy of focusing meta-training on the classifier alone.
Both simplify model optimization and finally improve the model performance.
By adapting the classifier to every query sample as in our full CWT model,
the performance can be further improved significantly. 
This validates the design of our CWT
for solving the intra-class variation challenge.
As shown in Figure~\ref{fig:visual},
the baseline fails to detect the airplane (1$^{st}$ row) 
and the person (2$^{nd}$ row) due to the lack of query adaptation. These failure cases are rectified once query-image adaptation is in place using the proposed CWT. 




\begin{table}[]
    \centering
    \begin{tabular}{c|ccccc}
    \hline
         \multirow{2}{*}{Attend to} & \multicolumn{5}{c}{1-shot}  \\
         \cline{2-6}
          & s-0 & s-1 & s-2 & s-3 & Mean \\
         \hline
         Support image & 20.1 & 18.9 & 21.9 & 11.9 & 18.2  \\
         Query image & \textbf{32.2} & \textbf{36.0} & \textbf{31.6} & \textbf{31.6} & \textbf{32.9} \\
         \hline
    \end{tabular}
    \vspace{2pt}
    \caption{Effect of intra-class variation.
    Backbone: ResNet-50,
    Dataset: COCO-20$^i$.}
    \vspace{-0.4cm}
    \label{tab:support}
\end{table}


\vspace{-0.3cm}

\subsubsection{Importance of query adaptation}
Recall that our CWT is designed primarily to address the intra-class variation problem by adapting the classifier weights initialized on the support set to each query image.
To further validate this module,
we contrast with another transformer design without involving
the query image whilst still remaining the attention learning ability.
Concretely, we set the support feature as the \texttt{Key} and \texttt{Value} inputs of the transformer, instead of the query feature.
This ignores the intra-class variation issue in design.
The results in Table~\ref{tab:support} show that 
without conditioning on query image, the segmentation performance degrades drastically. 
This indicates the essential importance of resolving the intra-class
variation problem and the clear effectiveness of the proposed design.



\begin{figure}
    \centering
    \includegraphics[width=0.45\textwidth]{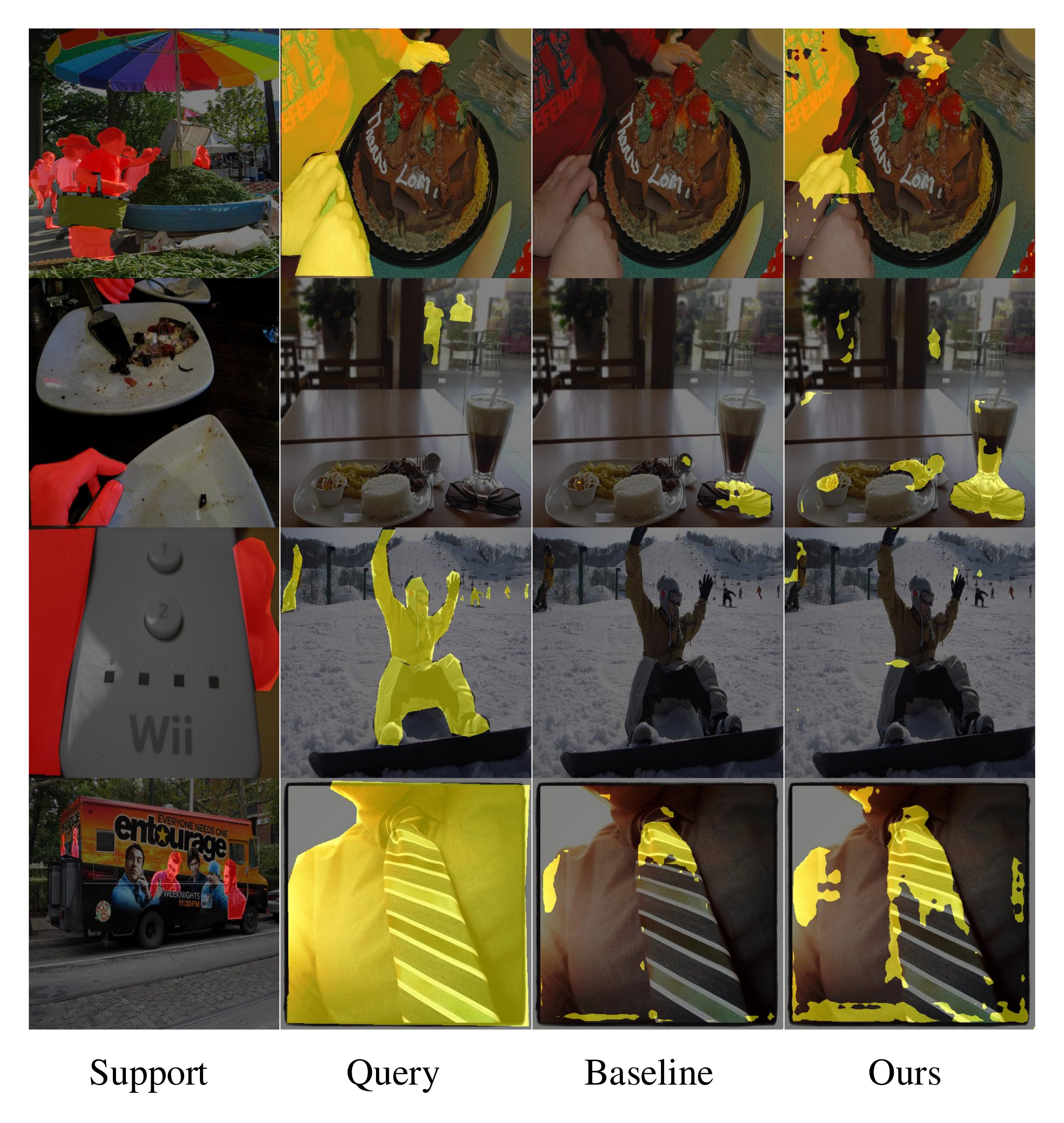}
    \caption{Failure cases.
    Backbone: ResNet-50,
    Dataset: COCO-20$^i$,
    1-shot setting.
    }
    \label{fig:failure}
\end{figure}

\vspace{-0.3cm}

\subsubsection{Failure Cases}
Beyond the numerical evaluations as shown above,
we further study failure cases in Figure~\ref{fig:failure}.
This can give us some insights and potentially inspiring directions for the future investigation. Overall we observe that 
our model fails when the target instances with extreme object appearance changes exist between the support and query images.
For instance, the support image presents only the hands of a person whilst the query image covers a whole person body (see the 2$^{nd}$ and 3$^{rd}$ rows).
In contrast, the failure in the 1$^{st}$ and 4$^{th}$ rows would be caused mainly by extreme viewpoint differences.
How to deal with such appearance variation requires better modeling of changes in views, pose and occlusion. 


\vspace{-0.2cm}

\section{Conclusion}
We have presented a novel few-shot segmentation learning method with meta-learning. Our method differs significantly from existing ones in that we only meta-learn the classifier part of a complex segmentation model whilst freezing the pre-trained encoder and decoder parts. 
To address the intra-class variation issue, we further propose a Classifier Weight Transformer (CWT) for adapting the classifier's weights, which is first initialized on a support set, to every query image.
Extensive experiments verify the performance superiority of 
our proposed method over the existing state-of-the-art few-shot segmentation methods on two standard benchmarks.
Besides, we investigate a more challenging and realistic setting -- \textit{cross-domain few-shot segmentation},
and show the advantages of the proposed method.

{\small
\bibliographystyle{ieee_fullname}
\bibliography{egbib}
}

\end{document}